\documentclass{article} 
\usepackage{iclr2017_workshop,times}
\usepackage{url}

\usepackage{amsmath} 
\usepackage{amssymb}  
\usepackage{amsfonts}  
\usepackage{graphicx} 
\usepackage[font=small,labelfont=bf,tableposition=top]{caption}
\usepackage{booktabs}
\usepackage{multicol}
\usepackage{subfigure}
\usepackage{cuted}
\usepackage{microtype}
\usepackage{fancyhdr}
\usepackage[colorlinks=true,
        linkcolor=black,   
        citecolor=black,   
        filecolor=black,
        urlcolor=black,
        bookmarks=true,
        bookmarksopen=true,
        bookmarksopenlevel=3,
        plainpages=false,
        pdfpagelabels=true]{hyperref}
\usepackage{microtype}
\usepackage{cite}
\usepackage{multirow}

\usepackage{titlesec}
\titlespacing*{\paragraph}{0pt}{0.1\baselineskip}{\baselineskip}

\newcommand{\vect}[1]{\mathbf{ #1}}

\newcommand{\vN}{\vect{N}}

\newcommand{\vx}{\vect{x}}

\newcommand{\cA}{\mathcal{A}}

\newcommand{\cO}{\mathcal{O}}

\titlespacing\section{0pt}{4pt plus 2pt minus 2pt}{4pt plus 2pt minus 2pt}   

\title{Episode-Based Active Learning\\with Bayesian Neural Networks}

\author{Feras Dayoub, Niko S\"underhauf, Peter Corke\\
Australian Centre for Robotic Vision\\
Queensland University of Technology (QUT)\\
Brisbane, 4000 QLD, Australia\\
\texttt{\{feras.dayoub, niko.suenderhauf, peter.corke\}@qut.edu.au} \\
}

\begin{document}

\maketitle
\pagestyle{empty}

\begin{abstract}
We investigate different strategies for active learning with Bayesian deep neural networks. We focus our analysis on scenarios where new, unlabeled data is obtained \emph{episodically}, such as commonly encountered in mobile robotics applications. An evaluation of different strategies for acquisition, updating, and final training on the CIFAR-10 dataset shows that incremental network updates with final training on the accumulated acquisition set are essential for best performance, while limiting the amount of required human labeling labor.
\end{abstract}

\section{Introduction}
Obtaining labeled training data is a big challenge for a robotic scene understanding system that is pre-trained on a dataset, but then has to adopt to a real world deployment environment. Active learning, \citet{cohn1996active}, helps to minimize the necessary human labeling labor by acquiring only the most informative samples from a pool of available images.

In the context of mobile robotics, active learning happens \emph{episodically}. Starting with an initial classifier, the robot encounters a stream of images while performing its mission. After a certain time (an \emph{episode}), an acquisition function determines the most informative encountered images, and a human can be asked to provide ground truth labels. The initial network is then updated using this set of acquired images. The process repeats as new, previously unseen images are encountered during the next episode.

In this paper we investigate different strategies of performing active learning in this episode-based scenario. We aid the acquisition function by using Bayesian deep networks as classifiers. Our goal is to enable a mobile robot to adopt its perception system to its deployment environment with as little human help and interaction as possible.

\section{Episode-based Active Learning}
\paragraph*{Notation}
We use $\vN_t$ to represent a network obtained after episode $t$ and write $\vN_t = \vN_{t-1} \otimes \{\cA\}$ to express fine-tuning network $\vN_{t-1}$ with the data set $\{\cA\}$ to obtain $\vN_t$.

\paragraph*{Problem Definition}
We define the problem of episode-based active learning as follows: Start with an initial, pre-trained network $\vN_0$. For episodes $t=1\dots k$ perform the following steps: (1) present an episode of $n$ previously unseen and unknown images to the network $\vN_{t-1}$, and obtain classification results $p(\vx_i)$, $i=1\dots n$; (2) based on $p(\vx_i)$, use an acquisition function to determine the set $\cA_t$ of most informative images and ask an oracle (a human) for their ground truth labels; (3) update the network $\vN_{t-1}$ by fine tuning with the acquired dataset $\cA_t$. After stopping the active learning process after $k$ episodes (e.g. based on the the number of acquired images per episode, or simply based on the number of passed episodes $k$), a final training step might be performed to obtain the final network $\vN_f$.

Different strategies are available for updating the network, for the final training, and for the acquisition function. Table~\ref{tab:scenarios} lists the different combinations of strategies we investigated and evaluated in this paper.

\paragraph*{Network Update}
Updating the network after every episode can be done using only the newly acquired set $\cA_t$ from that episode,  or by using the full data $\cup_{i=1}^{t} \cA_i$ that has been acquired so far. We will evaluate four different methods for updating the network $\vN_{t-1}$: incremental fine-tuning using only the most recently acquired images: $\vN_t = \vN_{t-1} \otimes \{\cA_t\}$; incremental updating using the growing set of all acquired images so far: $\vN_t = \vN_{t-1} \otimes \{\cup_{i=1}^{t} \cA_i \}$; and fine-tuning on the initial network: $\vN_t = \vN_{0} \otimes \{\cup_{i=1}^{t} \cA_i \}$, and $\vN_t = \vN_0 \otimes \{\cA_t\}$.

\paragraph*{Acquisition Function}
We apply a maximum entropy acquisition strategy that selects the images with the highest classification uncertainty from an episode.
\citet{gal2017bayesian} showed this strategy to be competitive to more complex acquisition schemes. For every image $\vx_i$ in the current episode, we obtain the class probability distribution $p(\vx_i)$ from the current network $\vN_t$. An image is acquired if $H(p(\vx_i)) > \theta$, where $H(\cdot)$ denotes the entropy and $\theta$ is a threshold parameter.

In our evaluation, the true class probability distribution $p(\vx_i)$ is approximated by a Bayesian neural network approach proposed by \citet{gal2015bayesian}. We enable Dropout during test time, pass every image through the network 64 times, and average over the obtained 64 distributions.

\paragraph*{Final Training}
The final network after $k$ episodes can be obtained either by simply using the network $\vN_k$ from the last update step, or by using all acquired images to perform an additional fine-tuning: $\vN_f = \vN_{k} \otimes \{\cup_{i=1}^{k} \cA_i \}$.

\begin{table}[]
    \centering
    \begin{tabular}{@{}lllccl@{}}
         \toprule
           &    &  &  & Used Trai- & Relative \\
         Strat. & Network Update & Final Training & Accur. & ning Set & Efficiency  \\
         \midrule
         1 & $\vN_t = \vN_{t-1} \otimes \{\cA_t\}$ & $\vN_f = \vN_{k} \otimes \{\cup_{i=1}^{k} \cA_i \}$ & \textbf{0.810} & 74\% & 1.36 \\
         2 & $\vN_t = \vN_{t-1} \otimes \{\cup_{i=1}^{t} \cA_i \}$  &  $\vN_f = \vN_k$
         & 0.808 & 70\% & \textbf{1.43} \\
         3 & $\vN_t = \vN_{t-1} \otimes \{\cA_t\}$ &  $\vN_f = \vN_{k}$
         & 0.767 & 73\% & 1.31  \\
         4 & $\vN_t = \vN_{0} \otimes \{\cA_t \}$ & $\vN_f = \vN_{0} \otimes \{\cup_{i=1}^{k} \cA_i \}$ & 0.793 & 82\% & 1.20 \\
         5 & $\vN_t = \vN_{0} \otimes \{\cup_{i=1}^{t} \cA_i \}$ & $\vN_f = \vN_{0} \otimes \{\cup_{i=1}^{k} \cA_i \}$  & 0.787 & 73\% & 1.34 \\
         \midrule
         6 & \multicolumn{2}{l}{regular training on the full training set}  & 0.805 & 100\% & 1.0\\
         7 & \multicolumn{2}{l}{regular training on random 74\% of the full training set}  & 0.781 & 74\% & 1.31\\
        \bottomrule
    \end{tabular}
    \caption{Different active learning strategies evaluated in this paper. Accuracy on the test set, the size of the used training set and the relative efficiency are reported for the final network $N_f$ after all 9 episodes.}
    \label{tab:scenarios}
\end{table}

\section{Evaluation on CIFAR-10}

\paragraph*{Dataset} We evaluate the 6 different strategies in Table~\ref{tab:scenarios} using the CIFAR-10 dataset. This standard dataset consists of 32$\times$32 pixel RGB images of ten classes and is split into 50,000 images for training and 10,000 for testing. We divide the training set into 10 splits of 5k images each. The first split is used to train the initial model $\vN_0$ that is shared between strategies. The rest of the non-overlapping splits is used as 9 episodes. All the testing is done on the 10k test images.

\paragraph*{Network Architecture}
We use a simple network consisting of 4 convolution layers (2$\times$32 and 2$\times$64 channels, convolutions are all 3$\times$3) followed by one fully connected layer (512 units) and a 10-fold Softmax layer. Dropout ($p=0.5$) is used after every layer, and max-pooling layers after conv layers 2 and 4. The network is trained with Adam,  \citet{Diederik15}, and early stopping.

\paragraph*{Evaluation Protocol}
For every strategy in Table~\ref{tab:scenarios} we evaluate the test set accuracy of the final network that would result from stopping the active learning after every episode.
All experiments are performed 10 times and the average accuracy and acquired number of training images are reported. To compare the performance of different strategies, we furthermore define an efficiency score as $\xi = \frac{\text{test accuracy}}{\text{fraction of used training set}}$. The efficiency of a network is high when it uses less training samples to gain a higher accuracy on the test set. Furthermore the \emph{relative efficiency} is defined as $\frac{\xi}{\xi_F}$, where $\xi_F$ is the efficiency of the network trained on the full training set.  We empirically set the acquisition threshold to $\theta=0.8$, but also analysed the influence of this parameter.

\begin{figure}[t]
\centering
  \centering
  \includegraphics[width=0.49\linewidth]{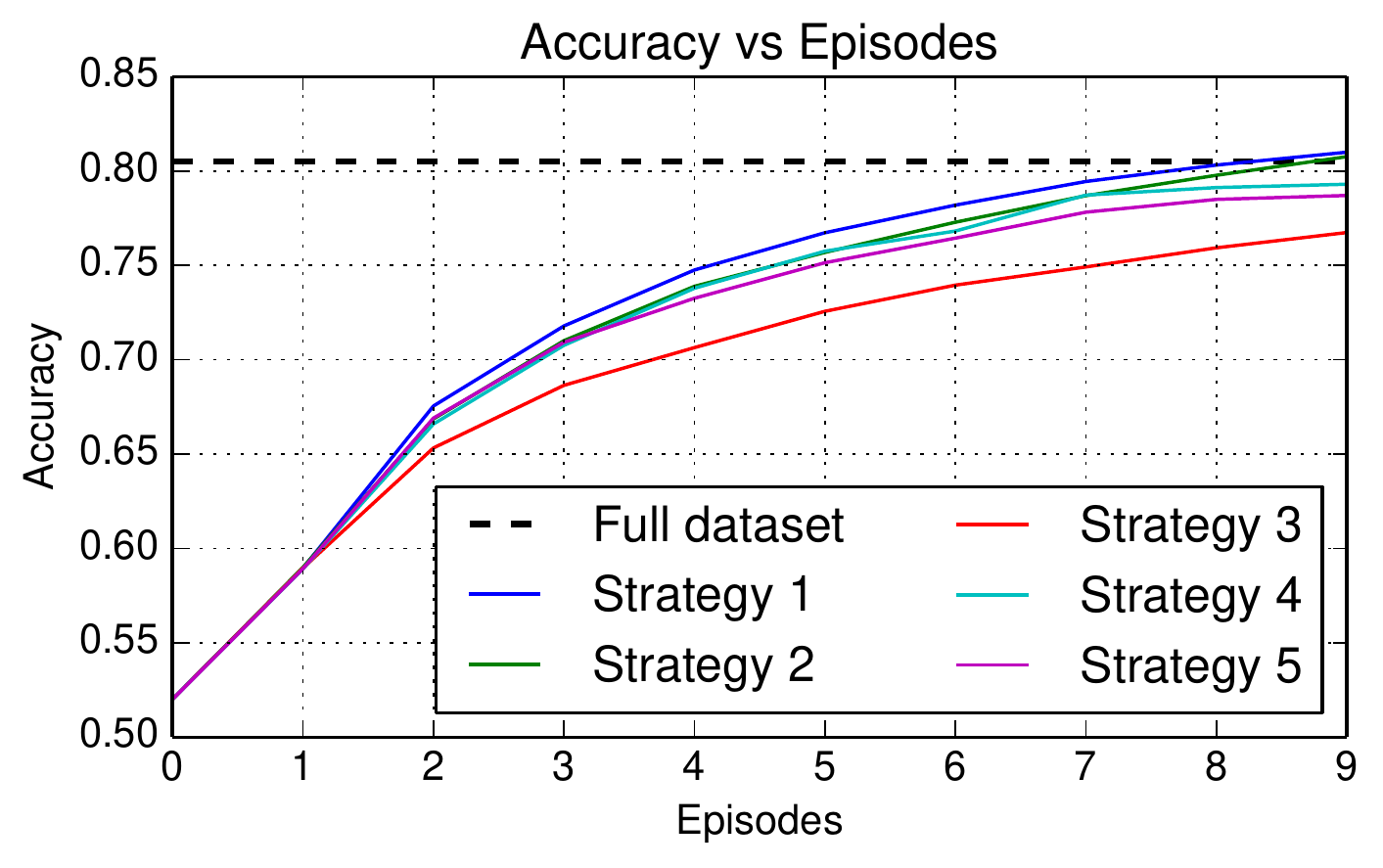}
  \hfill
  \includegraphics[width=0.49\linewidth]{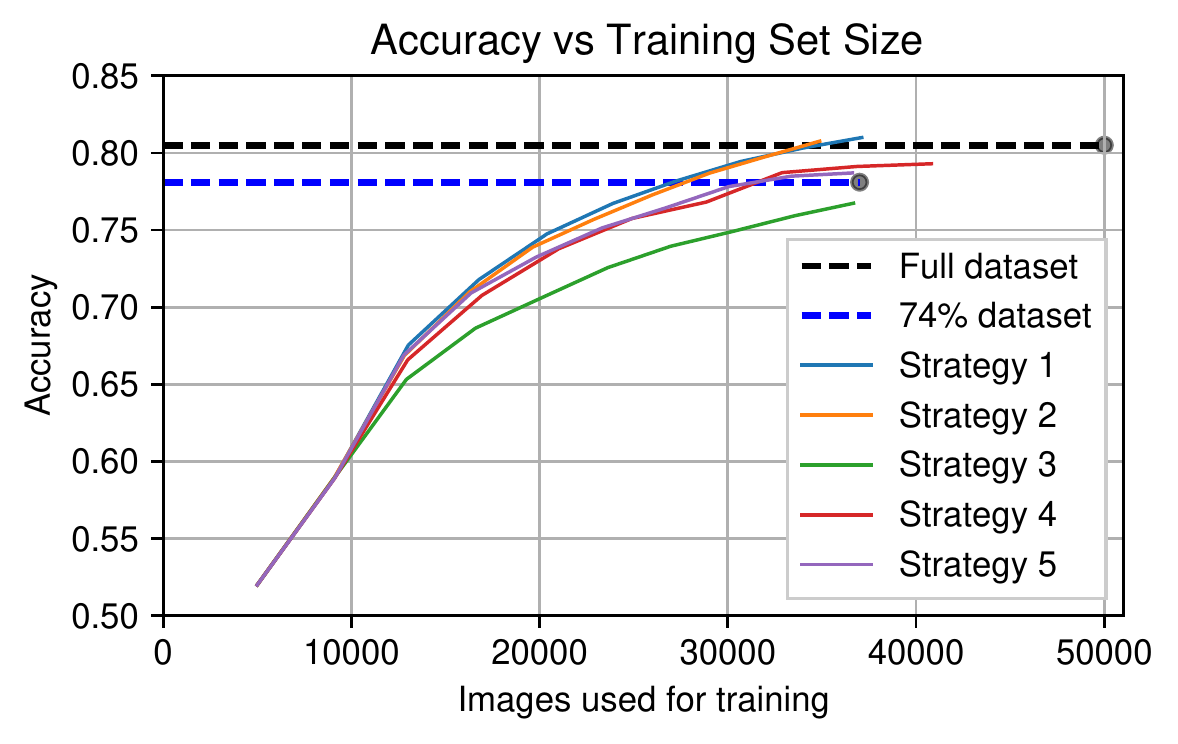}
  \caption{Test set accuracy for various active learning strategies. Incremental fine-tuning with final training on the accumulated acquired images (strategies 1 and 2) outperforms other variants and even reaches slightly better performance than the baseline model that was trained on the full dataset. Plots averaged over 10 trials.}
  \label{fig:results1}
\end{figure}

\paragraph*{Results}
Our evaluation showed that strategies 1 and 2 which use incremental fine-tuning during the update step with a final training step on the accumulated acquired images outperform the other strategies. In particular, strategies 4 and 5 that update based on the initial network $\vN_0$ perform worse. This is illustrated in the plots of Fig. \ref{fig:results1}, where strategies 1 and 2 learn faster (i.e. gain more accuracy per episode) and more efficient (i.e. gain more accuracy per acquired images). Both strategies even outperform the baseline model (training on the full dataset) by a small margin. Strategy 2 is more data-efficient, using 4\% less training data while only sacrificing 0.2\% of test accuracy. This comes at an increased computational cost for the update step, which is in $\cO(n)$ instead of $\cO(1)$ due to the growing update fine-tuning dataset $\cup_{i=1}^{t} \cA_i$. Training on randomly selected 74\% of the training data results in a significantly worse performance, which underlines the efficacy of the proposed active learning strategies (dashed blue line in Fig. \ref{fig:results1} (right)).

The differences between strategy 3 and the first two strategies illustrate the importance of using the accumulated acquired images $\cup_{i=1}^{k} \cA_i$ during updating or at least during final training. We assume this prevents the \emph{forgetting} that seems to occur when using only the most recent acquired images.

\begin{figure}[t]
\centering
  \centering
  \includegraphics[width=0.49\linewidth]{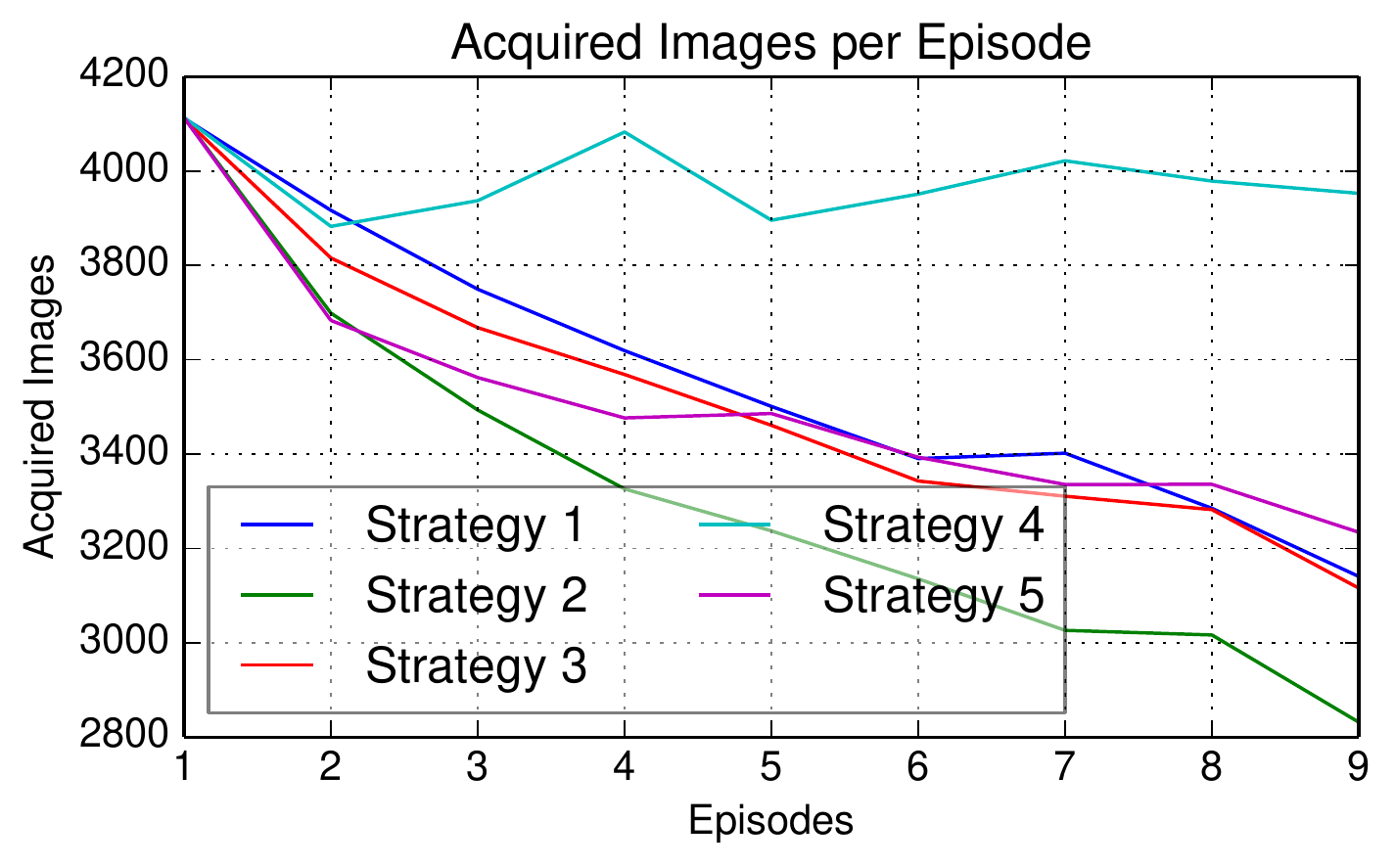}
  \hfill
  \includegraphics[width=0.49\linewidth]{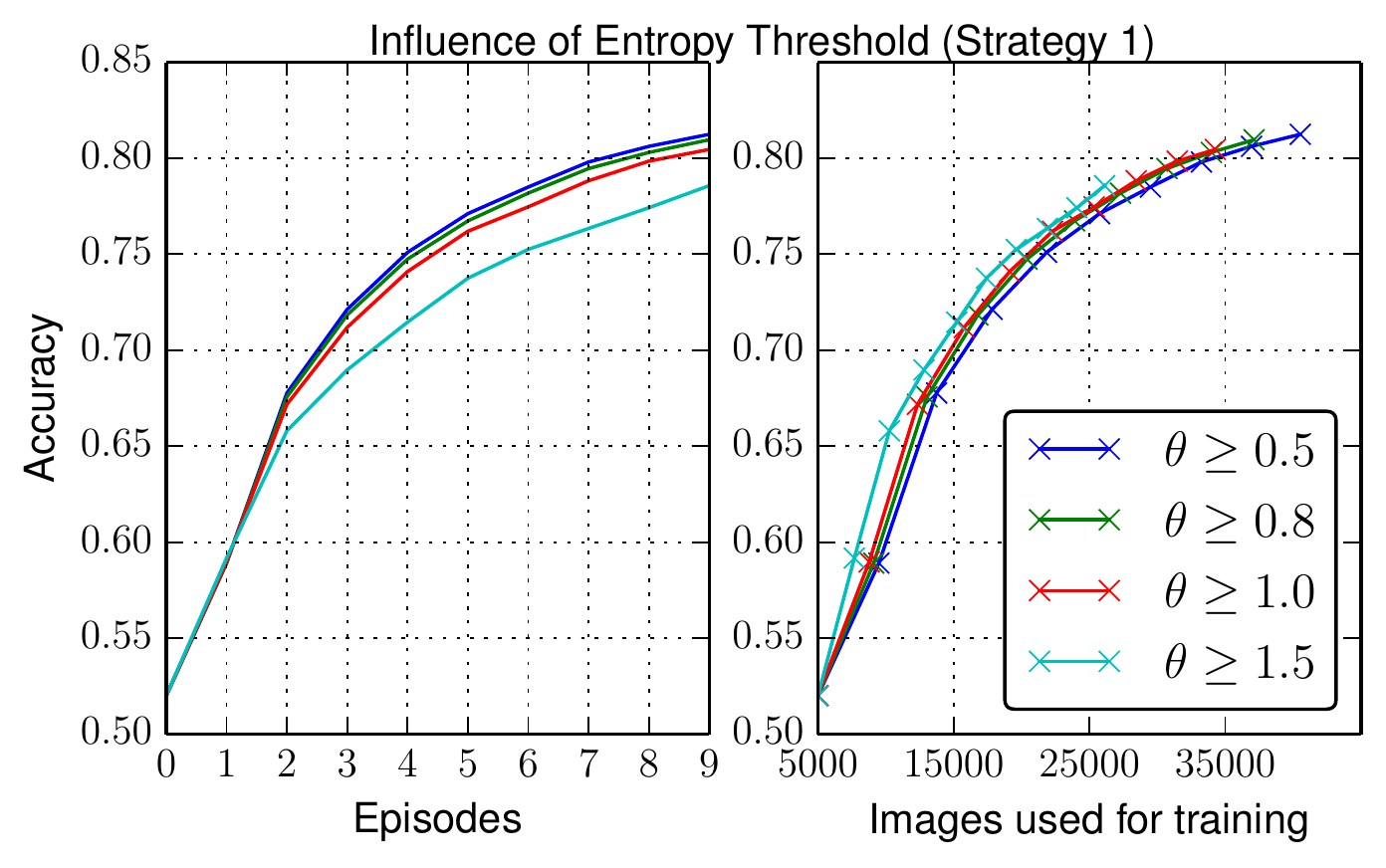}
  \caption{(left) The acquisition function and network update strategy determine the data-efficiency: Strategy 2 acquires the least number of images, but outperforms even the baseline model in terms of test set accuracy. (right) Influence of the selection threshold $\theta$ in the max entropy acquisition function evaluated on strategy 1.
  }
  \label{fig:results2}
\end{figure}

Fig. \ref{fig:results2}(left) compares the number of images acquired by every strategy over the 9 episodes. Strategy 2 is very data-efficient, acquiring the least amount of images, while reaching the second highest accuracy. This is expressed in its relative efficiency score, that is the highest of all strategies. Notice how strategy 4 always acquires roughly the same amount of images. This is not surprising given its update strategy uses the initial network $\vN_0$ and only the most recent acquisition set $\cA_t$, thereby not learning more about the environment over time as the other strategies do.

Fig. \ref{fig:results2}(right) reveals the influence of the parameter $\theta$ used as a threshold in the maximum entropy acquisition function. As expected, a higher threshold acquires fewer images, which ultimately leads to a smaller training set and sacrifices accuracy.

\section{Conclusions}
Our evaluation found incremental update strategies with a final training step based on the accumulated acquired image set performed best for episode-based active learning with Bayesian neural networks.
In contrast to earlier work such as \citet{islam206master, gal2016workshop} that demonstrated active learning ideas with Bayesian networks in a pool-based setup on MNIST, we presented experiments on the more challenging CIFAR-10 dataset. Furthermore, we evaluated on an \emph{episode}-based scenario where the system does not get a chance to re-observe images it did not choose to acquire. This scenario is closer to the requirements encountered by a robot in reality.

\subsubsection*{Acknowledgments}
This research was conducted by the Australian Research Council Centre of Excellence for Robotic Vision (project number CE140100016).

\bibliography{refs.bib}
\bibliographystyle{iclr2017_workshop}

\end{document}